 \documentclass[tablecaption=bottom]{jmlr}% journal article article
\usepackage{booktabs}

\usepackage{wrapfig}

\usepackage[load-configurations=version-1]{siunitx} % newer version
 \usepackage[font=small]{caption}
\theorembodyfont{\upshape}
\theoremheaderfont{\scshape}
\theorempostheader{:}
\theoremsep{\newline}

\jmlrvolume{vol 144}
\jmlryear{2021}
\jmlrproceedings{PMLR}{Proceedings of Machine Learning Research}

\title[Optimal Cost Design for Model Predictive Control]{Optimal Cost Design for Model Predictive Control}

\newcommand{\utraj}{\ensuremath{\vec{u}}}

\usepackage[]{lineno}

\usepackage{xcolor}

\usepackage{bbold}

\newcommand{\E}{\mathop{\mathbb{E}}}

\author{%
\Name{Avik Jain} \Email{avikj@berkeley.edy}\\
 \Name{Lawrence Chan} \Email{chanlaw@berkeley.edu}\\
 \Name{Daniel S. Brown} \Email{dsbrown@berkeley.edu}\\
 \Name{Anca D. Dragan} \Email{anca@berkeley.edu}\\
 \addr InterACT Lab, University of California, Berkeley, USA}

\begin{document}
%\linenumbers
\maketitle

\begin{abstract}
Many robotics domains use some form of nonconvex model predictive control (MPC) for planning, which sets a reduced time horizon, performs trajectory optimization, and replans at every step. The actual task typically requires a much longer horizon than is computationally tractable, and is specified via a cost function that cumulates over that full horizon. For instance, an autonomous car may have a cost function that makes a desired trade-off between efficiency, safety, and obeying traffic laws.
%, all of which are part of its cost function for each step. 
%When we take a task specification and solve it via MPC, we simply accumulate the specified cost function over the short (MPC) planning horizon and implicitly assume the robot will follow the resulting short horizon trajectory, even though in reality it only follows the first step. 
In this work, we challenge the common assumption that the cost we optimize using MPC should be the same as the ground truth cost for the task (plus a terminal cost). 
MPC solvers can suffer from short planning horizons, local optima, incorrect dynamics models, and, importantly, fail to account for \emph{future replanning} ability. Thus, we propose that in many tasks it could be beneficial to purposefully choose a different cost function for MPC to optimize: one that results in the MPC \emph{rollout} having low ground truth cost, rather than the MPC \emph{planned trajectory}. We formalize this as an optimal cost design problem, and propose a zeroth-order optimization-based approach that enables us to design optimal costs for an MPC planning robot in continuous MDPs. We test our approach in an autonomous driving domain where we find costs different from the ground truth that implicitly compensate for replanning, short horizon, incorrect dynamics models, and local minima issues. As an example, the learned cost incentivizes MPC to delay its decision until later, implicitly accounting for the fact that it will get more information in the future and be able to make a better decision. Code and videos available at \url{https://sites.google.com/berkeley.edu/ocd-mpc/}.
\end{abstract}
\begin{keywords}
Model Predictive Control, Optimal Cost Design, Planning under Uncertainty
%Optimal Reward Design, Planning under Uncertainty, Autonomous Driving
\end{keywords}

\section{Introduction}\label{sec:introduction}
Robotics planning algorithms often sacrifice optimality for tractability. One of the most well known and widely used planning paradigms in robotics is nonconvex model predictive control (MPC)~\citep{lee2011model,garcia1989MPC,poignet2000nonlinear,kunhe2005mobile,zhang2016learning,ostafew2016learning,thananjeyan2020abc,sripathy2021dynamic}. MPC optimizes trajectories based on a ground-truth cost function, $C$, that describes the task. 
 %MPC is an efficient approach for optimizing trajectories in high-dimensional, continuous spaces.
 Rather than trying to find a globally optimal trajectory with respect to $C$, MPC uses four heuristics to enable tractable planning:  (1) \textit{Short Horizon}: MPC plans for a finite horizon which is typically much shorter than the full horizon of the problem, leading to potentially myopic behavior; this is only partially mediated by learning a terminal cost added to the final state of the trajectory \citep{zhong2013value,napat2020practical}; (2) \textit{Local Optimization}: MPC performs local trajectory optimization, which while efficient, often leads to only locally, but not globally optimal behavior;
 (3) \textit{Replanning}: MPC repeatedly replans a new trajectory following each observation and takes the first action from this plan; however, this can lead to suboptimal behavior since the generated plans do not take into account the fact that the robot will replan based on future observations; this is in contrast to contingency planning~\citep{hardy2013contingency} and full POMDP solvers~\citep{lovejoy1991survey,shani2013survey}, which explicitly account for the fact that the robot will gather new information and be able to replan in the future; (4) \textit{Approximate Dynamics Model: } MPC plans into the future using an approximate model of dynamics. Repeatedly replanning allows an agent the chance to recover; however, systematic errors in the dynamics can still lead to compounding errors in the actual executed trajectory. %\dbnote{Not sure if dynamics should be counted as an MPC heuristic (\#4) or just another aspect of MPC that can be alleviated by a surrogate cost...}
 %, ; this is only partially alleviated by approximations like contingency planning~\cite{hardy2013contingency}, which typically choose a single observation to branch on.%approaches, such as , that branch on possible future observations, 
 % \dbnote{TODO: write out a paragraph here following Anca's suggestions below and what she wrote in the abstract.}
% \adnote{see how i posed it in the abstract: there is a cost evaluated on full horizon, cumulatively per time step; naturally MPC optimizes that but only for a short horizon; we question this assumption; our idea is that we should lie to MPC about what we care about, so that when it suboptimally optimizes it, it leads to good behaviro for what we actually care about}
% \adnote{then give the intuition of WHY this is true, using e.g. a finite horizon and a replanning example}
% \adnote{one thing that will come up is why not just learn a terminal cost, so you need to say that while that alleviates the horizon issue, a) it's still not perfect because the terminal cost isnt perfect, leaving room for the cost to help; and b) there is also replanning and local optimization}
%MPC is often used to optimize the ground-truth cost function, $C$,that describes the task.% that would be given to an optimal planner. %MPC planners typically optimize this cost, but accumulated only over a shorter MPC planning horizon instead of the true horizon. 
%\begin{quote}

In this paper, we challenge the implicit assumption that MPC should optimize the true cost $C$, and instead propose that optimizing a surrogate cost $C'$ can actually improve performance of the resulting MPC rollout with respect to $C$.
    %Our insight is that this might not be the best choice: the cost that MPC should optimize need not be equal to the true cost $C$. 
%\end{quote}
Rather than focusing on making the MPC \emph{planned trajectory} have low true cost, we should focus on the MPC \emph{rollout} having low true cost. We observe that optimizing the true cost directly via MPC enables the former, but might not be an optimal choice for the latter.

\vspace{.1cm}
\noindent\textbf{Implicitly accounting for horizon issues:} As an example, consider an autonomous car (orange) driving in the scenario shown in \figureref{subfig:scenario_finite}. Here, the optimal behavior with respect to the robot's true cost is to switch lanes to maintain its speed. From the perspective of an MPC planner with a short planning horizon, the cost of switching lanes outweighs the benefit of maintaining its speed accumulated over the short horizon.
%, and so the robot car planning with MPC chooses instead to slow down. 
A learned surrogate cost function $C'$ incentivizes lane changing, thereby decreasing the true cumulative rollout cost.
Prior work seeks to address the issue of short horizon by learning a value function or terminal cost, evaluated at the final state of the MPC trajectory \citep{zhong2013value,lowrey2018plan,hoeller2020deep,napat2020practical}.
However, while prior work assumes you should optimize $C$ with an added terminal cost, we investigate replacing $C$ with a learned surrogate $C'$. The learned terminal cost will not be perfect (otherwise we would just act greedily with respect to it), so there is still headroom for a surrogate cost function $C'$ to improve performance even when we add a terminal cost. Our experimental results indicate that our proposed approach is complementary to learning a terminal cost: while learning a terminal cost results in better MPC performance by alleviating short planning horizon issues, using a learned terminal cost along with a learned surrogate cost function results in even better performance.
%We perform experiments with and without a learned terminal cost and find that because therlearned terminal costs are imperfect, which still gives headroom for a cost $C'$ to further improve performance. 
More importantly, a short horizon is not the only reason MPC can result in suboptimal behavior, MPC may also suffer from local optima and inaccurate dynamics models, and fails to account for future replanning. %\figureref{subfig:scenario_local} and \figureref{subfig:scenario_replanning} show examples of this, where using a different cost $C'$ helps compensate for these sources of suboptimality.  

\vspace{.1cm}
\noindent\textbf{Implicitly accounting for local optimization issues:} Consider the scenario in \figureref{subfig:scenario_local}. Here the optimal behavior with respect to the robot's true cost is to merge into the rightmost lane while avoiding collisions; however, there is a human car in the way. Because MPC only searches locally for a trajectory, it does not discover that a significant reduction in speed will let the human car pass and provide a collision free path to the right lane, and instead chooses to stay in the left lane. A learned surrogate cost function $C'$ smooths the optimization landscape and pushes the car out of the left lane, enabling MPC to more easily discover the globally optimal trajectory. 

\vspace{.1cm}
\noindent\textbf{Implicitly accounting for future replanning:} Even more interesting is the scenario shown in \figureref{subfig:scenario_replanning}, where the robot car is trying to travel as fast as possible while staying on the road and avoiding collisions; however, there is a human car driving slowly on the lane line in front of the robot, and the robot is uncertain about which lane the human will eventually merge into. Running MPC on the true cost function results in an overly conservative behavior that picks a lane to merge into and slows down to avoid a collision. This is because MPC plans ahead and sees that no matter which lane it picks, there is a chance that the human picks the same lane. By contrast, the learned surrogate cost function $C'$ actually \emph{encourages MPC to defer the decision of which lane to choose until later, when it will have more information about which lane the human driver chooses.} Thus, MPC with the learned surrogate cost function results in an emergent contingency planning behavior.
%, which allows the robot to achieve much better performance under the true objective by not reducing speed and delaying lane merging until the human reveals their lane preference.
%Another example is given in \figureref{fig:case_study_locality}, where the optimal behavior is to slow down and switch to the right lane without colliding into the human. Due to local optimization, an MPC planner with bad initial trajectories may not find this optimal trajectory. Similarly, in this second case, a shaped cost $C'$ that smooths out the optimization landscape can allow the MPC planner to more easily discover the optimal, lane-changing trajectory. 

\vspace{.1cm}
\noindent\textbf{Implicitly accounting for an approximate dynamics model:}
%An important reason for MPC replanning is because the planner uses an approximate dynamics model. Thus, MPC plans a trajectory using an inaccurate model of the dynamics. A learned surrogate cost function can implicitly compensates for dynamics mismatch, resulting in significant cost reductions when compared with vanilla MPC on the true cost. For example, 
Consider planning using a deterministic dynamics model, but where the true dynamics are subject to systematic noise, e.g., wind blowing across the highway. Learning a cost function can implicitly correct for the mismatch between the MPC dynamics model and the true dynamics by shaping the cost such that the car implicitly adjusts its steering to compensate for the wind.

%In the above scenarios, there exists a different cost function $C' \neq C$ such that the trajectory resulting from optimizing $C'$ with MPC has lower cost under $C$ than the trajectory that was optimized under $C$. 
\vspace{.1cm}
We demonstrate that we can address the above issues via  
%The task of finding an optimal $C'$ for a bounded planner to maximize a ground truth cost $C$ is known as 
\textit{optimal cost design}\footnote{This problem is usually called \textit{optimal reward design} in the RL setting, e.g., \cite{singh2009rewards}.}: given a ground-truth cost function $C$ and a robot planning algorithm, we find a surrogate cost function $C'$ such that optimizing for $C'$ results in robot trajectories with minimal cost under the true cost function $C$~\citep{singh2009rewards}.
Prior work on optimal cost design has focused on learning cost functions for bounded rational agents in the context of reinforcement learning, and approaches optimal cost design using Bellman backups, tree-based planning, or gradient descent on domains with discrete actions and often discrete state spaces~\citep{singh2009rewards,sorg2011optimal,sorg2010reward}. 
By contrast, we seek to solve the optimal cost design problems for non-linear dynamical systems with continuous states and actions that are optimized via nonconvex MPC. Other related work has investigated cost optimization for continuous control tasks. \cite{marco2016automatic} use Bayesian optimization to automatically tune the cost for an LQR controller. 
%We also apply zeroth-order cost optimization, but for the problem of MPC. 
\cite{napat2020practical} present a method for automatically shaping a sparse MPC cost function by learning a value function; however, while learning a value function can alleviate short horizon problems, it does not necessarily address issues that arise from local optimization, replanning, and approximate dynamics. 

We propose a zeroth-order optimization approach for optimal cost design for MPC. Given a set of initial conditions, we seek to learn a surrogate cost function $C'$ such that optimizing $C'$ results in MPC rollouts that have minimum cost under the true cost function $C$. We show in four case studies that this optimization is tractable and ameliorates the shortcomings of planning with MPC.
Our main contributions are as follows: (1) we highlight the implicit, but incorrect, assumption in MPC that we should always optimize the true cost function; (2) we analyze scenarios in which optimizing an alternate cost function via MPC yields lower cumulative costs than optimizing the true cost function; and (3) we explore the robustness of optimized surrogate cost functions to variations in scenario conditions.

\section{Optimal Cost Design for MPC}\label{sec:ocd-mpc}

We model our problem as a finite horizon, Markov decision process~\citep{puterman2014markov} with state $x \in \mathcal{X}$, controls $u\in \mathcal{U}$, transition function $x_{t+1} = f(x_t, u_t)$ where $t$ is the timestep, and cost $C_\theta$ parameterized by $\theta$. Given a starting state $x_0$, the goal is to optimize controls $\utraj = [u_0, \ldots, u_{T-1}]$ over horizon $T$ to minimize the cumulative cost according to $C_\theta$:
\begin{equation}
    \utraj^*(\theta) =  \operatorname*{argmin}_{\utraj} \sum_{t=0}^{T-1} C_\theta(x_t, u_t) \enspace.
    \label{eq:real_objective}
\end{equation}

\subsection{Model Predictive Control}
It is common in robotics to use online planning in the form of nonconvex model predictive control (MPC) \citep{garcia1989MPC,maciejowski2002predictive,lee2011model}. In MPC we usually do not have access to the true dynamics function $f$ and instead plan with respect to an approximate dynamics model $\hat{f}$. Furthermore, the robot may not get to directly observe everything about the state: even though we may model the world as an MDP, the real world is typically better modeled as a POMDP~\citep{smallwood1973optimal,lovejoy1991survey}. For example, in the autonomous vehicle setting considered in this paper, we often plan using MPC in the presence of other agents. When planning a driving trajectory, the robot needs to take into account the anticipated behavior of the other cars on the road. Thus, the robot needs a model of the other agents' behavior.  Following prior work by \cite{dorsa2017active}, we assume access to a model of human behavior parameterized by $\varphi$. Thus, the true state is a function of both the robot's controls and the controls of the other agent:
\[{x}^\varphi_{t+1} = f^\varphi(x_t, u_t) = f(x_t, u_t, u^\varphi_t) \]
However, in most cases, the robot does not know $\varphi$ and instead will maintain a belief over the expected behavior of the human. At time step $t$, MPC seeks to approximate the optimal controls $\utraj^*$ by optimizing the following objective:
\begin{equation}
    \utraj^{\rm MPC}_{t:t+K-1}(\theta) =  \operatorname*{argmin}_{\utraj_{t:t+K-1}} \E_\varphi \left [\sum_{i=0}^{K-1} C_\theta(\hat{x}^\varphi_{t + i}, u_{t+i}) \right ] \enspace.
    \label{eq:mpc_objective}
\end{equation}
where the expectation is taken over the robot's belief over $\varphi$ and $\hat{x}^\varphi_{t+1}$ is given by applying the robot's approximate dynamics model:
\begin{align*}
    \hat{x}^\varphi_{t+i+1} = \hat{f}^\varphi(x_t, u_t) = \hat{f}(\hat{x}_{t+i}, u_t, u^\varphi_t); ~\hat{x}^\varphi_t = x^\varphi_t,
\end{align*}
where $\utraj^{\rm MPC}_{t:t+K-1}(\theta)$ is found via trajectory optimization over the truncated horizon $K < T$.

% At time step $t$, MPC seeks to approximate the optimal controls $\utraj^*$ by optimizing the following objective:
% \begin{equation}
%     \utraj^{\rm MPC}_{t:t+K-1}(\theta) =  \operatorname*{argmin}_{\utraj_{t:t+K-1}} \sum_{i=0}^{K-1} C_\theta(x_{t + i}, u_{t+i}) \enspace.
%     \label{eq:mpc_objective}
% \end{equation}
% where $\utraj^{\rm MPC}_{t:t+K-1}(\theta)$ is found via trajectory optimization over the truncated horizon $K < T$. 
At time $t$, the robot executes its first planned action $u^{\rm MPC}_t(\theta) = \utraj^{\rm MPC}_{t:t+K-1}(\theta)[0]$, updates its belief over $\varphi$ based on the observed action of the other driver, and then replans $K$ steps into the future based on its next state $x_{t+1} = f^\varphi(x_t,u^{\rm MPC}_t) $. It is typically intractable to minimize \equationref{eq:mpc_objective} directly, so $\utraj^{\rm MPC}$ is found via local optimization, for example by performing $N$ steps of gradient descent:
\begin{equation}
    \utraj_{t:t+K}^{\rm MPC}(\theta) \gets  \utraj_{t:t+K}^{\rm MPC}(\theta) - \alpha \nabla_{\utraj} \E_\varphi \left [\sum_{i=0}^{K-1} C_\theta(\hat{x}_{t+i}^\varphi, u_{t + i}) \right ] \enspace. 
    \label{eq:mpc_local}
\end{equation}

The actual MPC rollout, consisting of visited states and executed controls, is as follows:
\begin{equation}
\xi_{\rm MPC} = \left((x^\varphi_0,u_{0}^{\rm MPC}(\theta)),(x^\varphi_1,u_{1}^{\rm MPC}(\theta)),  \ldots, (x^\varphi_{T-1}, u_{T-1}^{\rm MPC}(\theta))\right)
\end{equation}
where $u_{t}^{\rm MPC}(\theta) := \utraj_{t:t+K-1}^{\rm MPC}(\theta)[0]$ and 
%\begin{equation}
$x^\varphi_{t+1} = f^\varphi(x^\varphi_t, u^{\rm MPC}_{t}(\theta))$.
%= f(x_t, \utraj^{\rm MPC}_{t:t+K-1}(\theta)[0]).
%\end{equation}
Note that while MPC has the ability to anticipate future events via its approximate dynamics model $\hat{f}^\varphi$, and can take control actions accordingly by planning $K$ timesteps into the future, the plan is only executed for one step and then changes. Despite replanning at each timestep, MPC plans $K$ steps into the future assuming the entire trajectory will be executed.

% While any observation model $P$ may be used for belief updates, we will assume that human intent can be modeled as a cost function, $C^\varphi$ parameterized by $\varphi$. We follow the principle of maximum entropy \cite{ziebart2008maximum} and assume that actions with lower cost are exponentially more likely:
% \begin{equation}
%     P(u^\varphi_t \mid x_t, u_t^{\rm MPC}, \varphi) \propto \exp\left(\beta \sum_{i=t}^{T-1} C^\varphi(x_{i},u_{i}^{\rm MPC}, u_{i}^\varphi )\right),
% \end{equation}
% where the inverse temperature parameter $\beta$ modulates the suboptimality of the human. \dbnote{Not sure if we should formulate it this way or just say that during an episde we can get information about the human's type and then give more details for our replanning scenario. We don't really keep a Bayesian prior, but we kind of do assuming really high beta. But it's weird since we assume we know the point in the future when the human type will be known.}

\subsection{Problem Statement}
Given a robot that plans trajectories using MPC, a true cost function $C = C_\theta$, and a distribution of starting conditions $p_0$, we seek a surrogate cost function $C' = C_{\theta'}$, parameterized by $\theta'$, which is the solution to the following optimal cost design problem:
\begin{equation} \label{eq:ocd-mpc}
\theta' = \operatorname*{argmin}_{\theta'}   \E_{x_0\sim p_0, \varphi} \left [ \sum_{t=0}^{T-1} C_\theta(x^{\varphi}_t, u_t^{\rm MPC}(\theta')) \right ] \enspace,
\end{equation}
where $u_t^{\rm MPC}(\theta')$ denotes the first control planned using MPC by solving \equationref{eq:mpc_objective} under the surrogate cost parameters $\theta'$ and $x^\varphi_{t+1} = f^\varphi(x^\varphi_t,u_t^{\rm MPC}(\theta')) $. Note that, critically, \equationref{eq:ocd-mpc} seeks a surrogate cost function parameterized by $\theta'$ that performs well when the controls from a limited horizon MPC planner are evaluated over the longer, true horizon $T$. We call this problem Optimal Cost Design for Model Predictive Control or OCD-MPC.

\subsection{Finding a Surrogate Cost Function}

In theory, there always exists a surrogate cost function that allows MPC to recover the globally optimal trajectory.
% Trivially, assuming that the space of cost functions is sufficiently large, there exist costs that allow MPC to recover the optimal trajectory. 
For example, optimizing a cost that is defined as the L2 distance between the globally optimal controls and the MPC controls, $C'(x_t, u_t) = ||u^*_t - u^{\rm MPC}_t||^2$, will lead MPC to find the globally optimal controls, assuming that the local optimization converges. However, while this shows that an optimal surrogate cost function exists, such a cost function is impractical as it requires prior knowledge of the globally optimal controls. A more interesting question is: can we improve the behavior according to C with a more practical parameterization of the cost function space?

In this work, we seek to find a surrogate cost function $C'$ by optimizing \equationref{eq:ocd-mpc} directly. This optimization may be performed using a variety of methods such as Bayesian optimization \citep{snoek2012practical,shahriari2015taking} or evolutionary search methods~\citep{kennedy1995particle,pelikan2002survey}. We chose to use Covariance Matrix Adaptation Evolution Strategy (CMA-ES)~\citep{hansen2006cma}. CMA-ES is a stochastic zeroth-order (derivative-free) optimization algorithm that is well suited for non-convex optimization problems over continuous search spaces. CMA-ES also has very few hyperparameters that need manual tuning and enables easy parallelization for computational efficiency. %We initially tried using Bayesian optimization to find surrogate cost functions, but in preliminary experiments we found that CMA-ES was superior both in terms of run-time and performance of the optimized trajectories. 

Given a population of potential surrogate costs, we use MPC to generate rollouts under each proposed surrogate cost and then use the cumulative true cost under $C$ as the fitness.
%Though dynamics are deterministic in our experiments, 
We consider environments that have stochastic initializations in terms of car positions and velocities. Thus, we calculate the fitness of a proposal surrogate cost $C'$ as the average true trajectory cost when using MPC to optimize the surrogate cost proposal over $n$ i.i.d. samples from the initial state distribution $p_0$. This results in the following fitness function:
\begin{equation}
    \text{fitness}(\tilde{\theta}) = \frac{1}{n}\sum_{j=1}^n \sum_{t=0}^{T-1} C_\theta \left(x^{\varphi}_{j, t}, u_{j,t}^{\rm MPC}(\tilde{\theta}) \mid x^\varphi_{j,0}  \sim p_0 \right) \enspace,
    \label{eq:ocd-mpc}
\end{equation}
where $x^{\varphi}_{j, t}, u_{j,t}^{\rm MPC}$ are the $t$-th state-action pair from the $j$-th MPC rollout $\xi_j$. 

In \sectionref{sec:case_studies} we discuss four case studies that demonstrate the ability of OCD-MPC to optimize a surrogate cost function for a single initial state ($n=1$) sampled from $p_0$ and used at both training and test time. 
These experiments allow us to better analyze the qualitative and quantitative performance of OCD-MPC. In \sectionref{sec:generalization}, we explicitly study the generalization capability of OCD-MPC, where we optimize a surrogate cost funtion $C'$ under varying numbers of initial states $n$ and then evaluate the learned cost function on a larger held-out set of test initializations drawn from $p_0$.
%the same initial state distribution.

\section{Experiments}%\label{sec:experimental-design}
\label{sec:case_studies}
In this section, we empirically test whether OCD-MPC can learn surrogate cost functions that implicitly compensate for issues due to short horizons, local optima, failing to account for future replanning, and planning using an approximate dynamics model.  

\vspace{.1cm}
\noindent\textbf{Driving Simulator and Implementation Details:}
% \noindent\textbf{Environment.} 
We use an autonomous driving simulator based on \cite{sadigh2016planning, dorsa2017active}.\footnote{Code and videos available at: \url{https://sites.google.com/berkeley.edu/ocd-mpc/}}%\url{https://github.com/avikj/l4dc-mpc-ocd}} 
%In the next section we will consider the ramifications of OCD-MPD in several specific autonomous driving scenarios. Here we describe the particulars of our experimental setup.\footnote{For additional implementation details, consult our code at: \url{https://github.com/avikj/l4dc-mpc-ocd}}
% \subsection{Autonomous Driving Domain}
% Following \cite{sadigh2016planning, dorsa2017active}, we model the dynamics of cars using a point-mass model. The state of each car is a 4-dimensional vector  $x = [ a ~ b ~h ~v]$, where $a, b$ are the Cartesian coordinates of the car, $h$ is the heading, and $v$ is the speed. The control input for the car is a two dimensional vector $u = [u_1 ~ u_2]$, where $u_1$ is the steering input and $u_2$ is the acceleration. We also include a friction coefficient $\mu$. The dynamics model of the vehicle is:
% \begin{equation}
% [\dot{a} ~ \dot{b} ~\dot{h} ~\dot{v}] = [v\cdot\cos{h} ~~~~ v \cdot\sin{h} ~~~~ v \cdot u_1 ~~~~u_2 - \mu\cdot v].
% \end{equation}
% %which we discretize along the time dimension.
% %For ease of simulation, we discretized the simulation along the time dimension using the following dynamics:
% %\[\Delta{\state}_t  = [\bar v_t\cdot\cos{h_t} \cdot dt ~~~~ \bar v_t \cdot\sin{h_t} \cdot dt ~~~~ \bar v_t \cdot u_1 \cdot dt ~~~~ (u_2 - \alpha\cdot v_t)\cdot dt], \] 
% %where $ \Delta{\state}_t = \state_{t+1} - \state_t $ and $\bar v_t = v_t + 0.5 \cdot u_2 \cdot dt$. In our experiments, we used $dt = 0.1s$.
The robot's true cost function $C_\theta$ is modeled as a linear function of the following continuous features: the squared difference between its current forward velocity and a target velocity, %$f_{\textrm{speed}}(x_t, u_t) = (v_t \cdot \sin~h - v_{\textrm{target}})^2$; 
%a human car collision spike,
% \[f_{\textrm{collision}}(x_t, u_t) = \exp\left (-\frac{1}{1-\delta_x^2} \right) \ind_{|\delta_x| < 1} \exp\left (-\frac{1}{1-\delta_y^2} \right) \ind_{|\delta_y| < 1}\]
a human car collision spike,
a smoothed indicator function for grass or road,
distance from the center of the closest lane, and the distances from the center of each lane.\footnote{We implemented these features following \cite{sadigh2016planning}, except we replace Gaussian functions with smooth bump functions and logistic functions with smooth step functions.}
% \vspace{.2cm}
%\subsection{Implementation Details}
% \noindent\textbf{MPC:} 
MPC is run with a receding horizon of $K=5$. The true horizon is $T=15$ in Scenarios 1 and 2 and $T=20$ in Scenario 3. Unless otherwise specified, the planner optimizes controls with control initializations which go straight, steer right, and steer left.  MPC uses 100 steps of gradient descent for each initialization, and returns the plan with the lowest cost. We normalize cost function weights to have unit $l2$-norm.
%so the effective learning rate of the trajectory optimizer is constant. 
% \vspace{.2cm}
%\subsection{Optimization Details}
%We use the entropy search method of Bayesian optimization to search over surrogate cost functions. We use the Python RoBO library \cite{klein-bayesopt17} for implementing Bayesian Optimization. 
% \noindent\textbf{CMA-ES:} 
For CMA-ES, we used the PyCMA implementation~\citep{hansen2019pycma}. We set $\sigma_0=0.05$, and use the default increasing population schedule. We stop CMA-ES after evaluating 85 candidate cost functions and return the cost function weights with the highest fitness. We also compare CMA-ES to a random search baseline, which searches for a high-fitness surrogate cost function by uniformly sampling 85 unit-norm weight vectors.
% for cost functions, and returns the one which performed the best under the samples we use for evaluation.
%\section{Case Studies}
%We first examine the performance of \algname in three case studies which showcase some of the suboptimalities that can result when running MPC, and the benefits of optimal cost design in these situations. In particular, we examine how OCD-MPC can overcome suboptimal behavior that can result from (1) a short horizon, (2) local optimization, and (3) trajectory optimization that does not explicitly account for future replanning.  % and highlight how a learned cost function causes MPC to discover qualitatively better behavior with higher cost. 

\begin{figure}[t]
     \subfigure[\footnotesize{Scenario 1: Short Horizon}]
    {\label{subfig:scenario_finite}%
      \includegraphics[width=0.29 \textwidth]{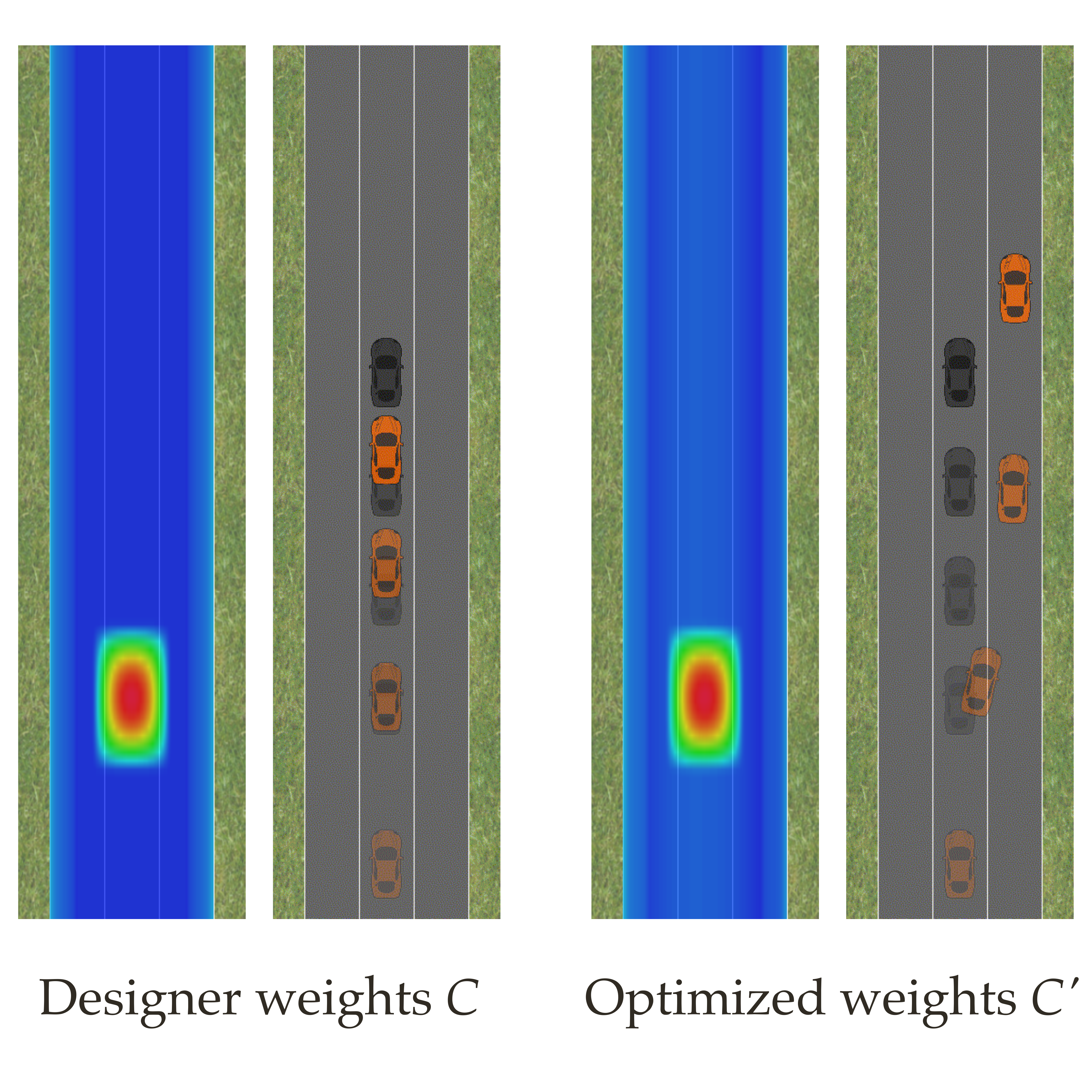}}%
    \hspace{0.01\textwidth}%
    \hfill
    \subfigure[\footnotesize Scenario 2: Local Opt.]
    {\label{subfig:scenario_local}%
      \includegraphics[width=0.29 \textwidth]{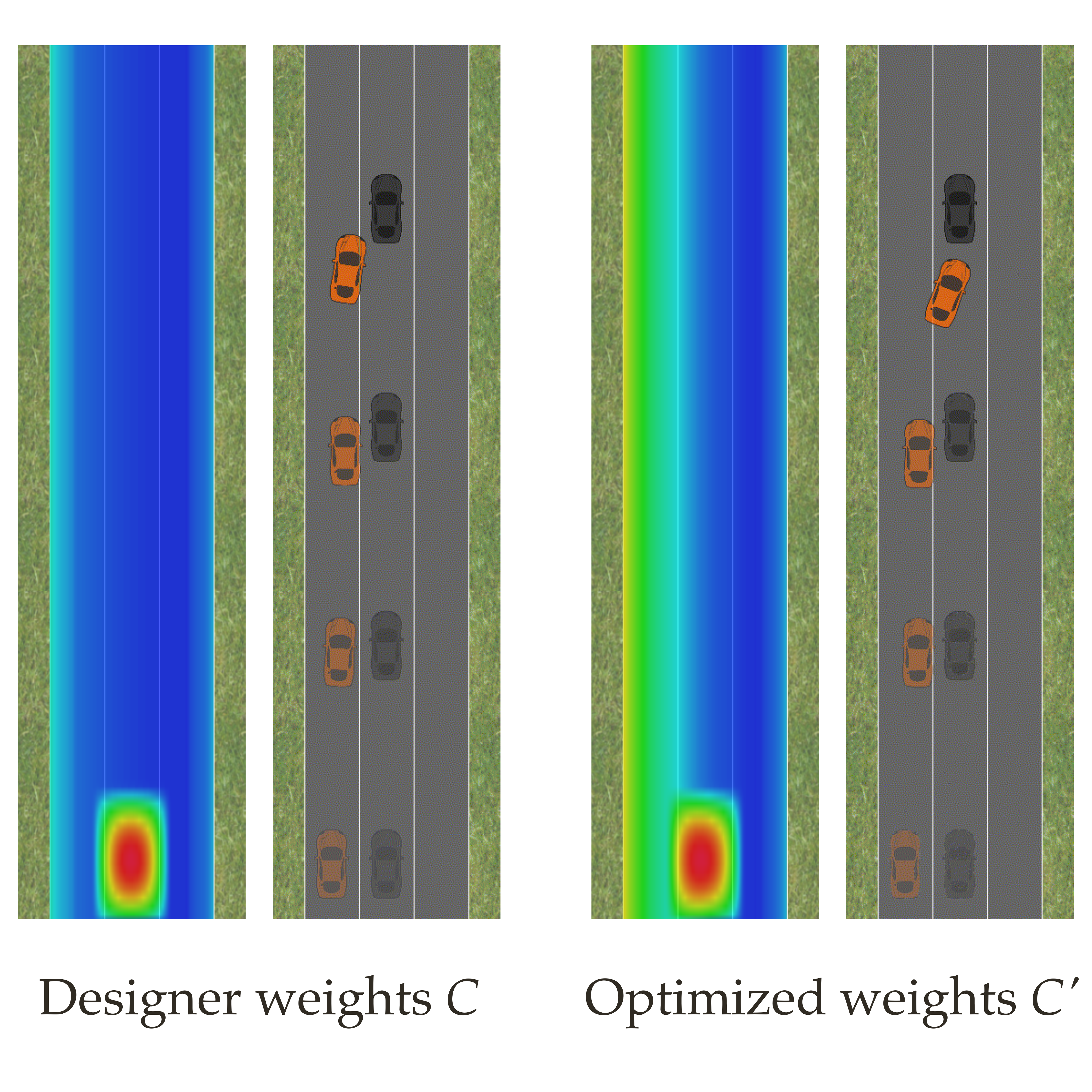}}%
    \hspace{0.01\textwidth}%
    \hfill
    \subfigure[\footnotesize Scenario 3: Replanning]
    {\label{subfig:scenario_replanning}%
      \includegraphics[width=0.39 \textwidth]{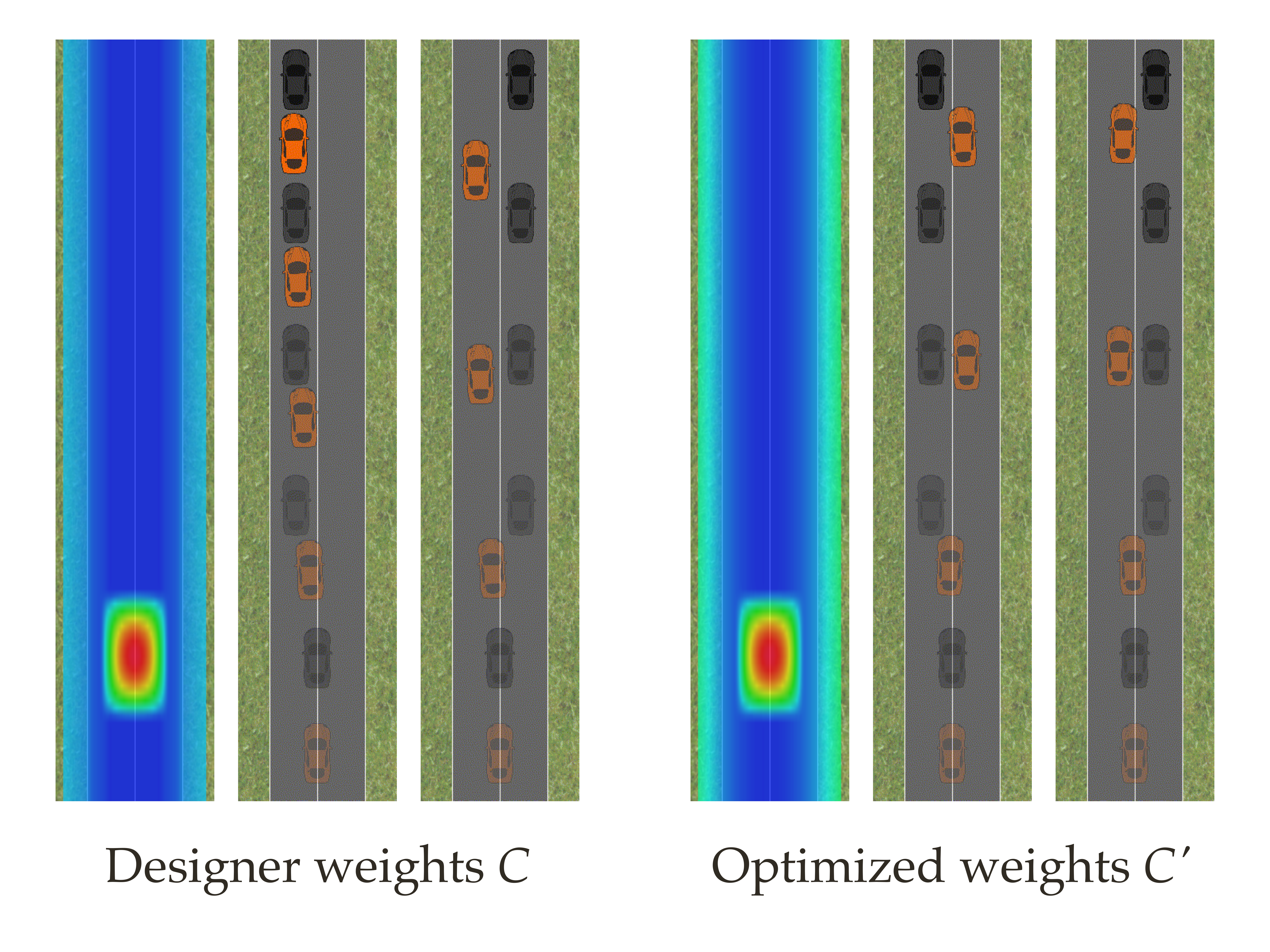}}%
     
        \caption{(a)-(c) Qualitative comparisons over three scenarios where planning with MPC under the true cost function $C$ results in suboptimal performance, but where these suboptimalities can be aleviated via %optimal cost design of 
  a surrogate cost function $C'$, which when optimized via MPC results in better performance under the true cost function $C$. The MPC robot is colored orange. See \url{https://sites.google.com/berkeley.edu/ocd-mpc/} for videos of the different behaviors. %(d) Quantitative comparison of performances.
  \vspace{-10pt}
  } 
        \label{fig:case_studies}
\end{figure}

\begin{figure}
    \centering
    \subfigure[\small No dynamics mismatch]{
    \label{fig:delta_perf_scenarios}
    \includegraphics[width=0.46\textwidth]{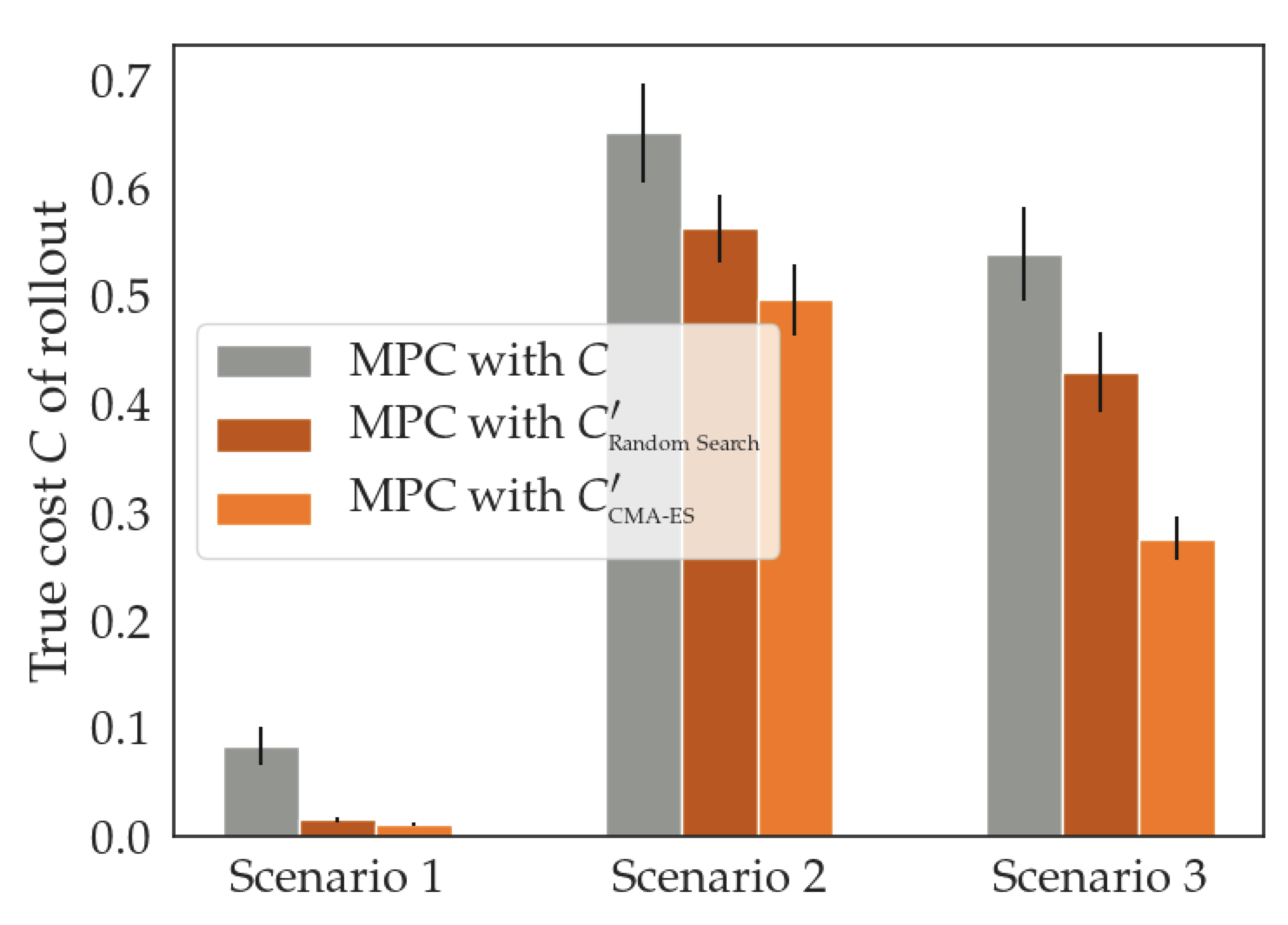}
    }
    \subfigure[\small Dynamics mismatch]{
    \label{fig:delta_perf_scenarios_dynamics_mismatch}
    \includegraphics[width=0.44\textwidth]{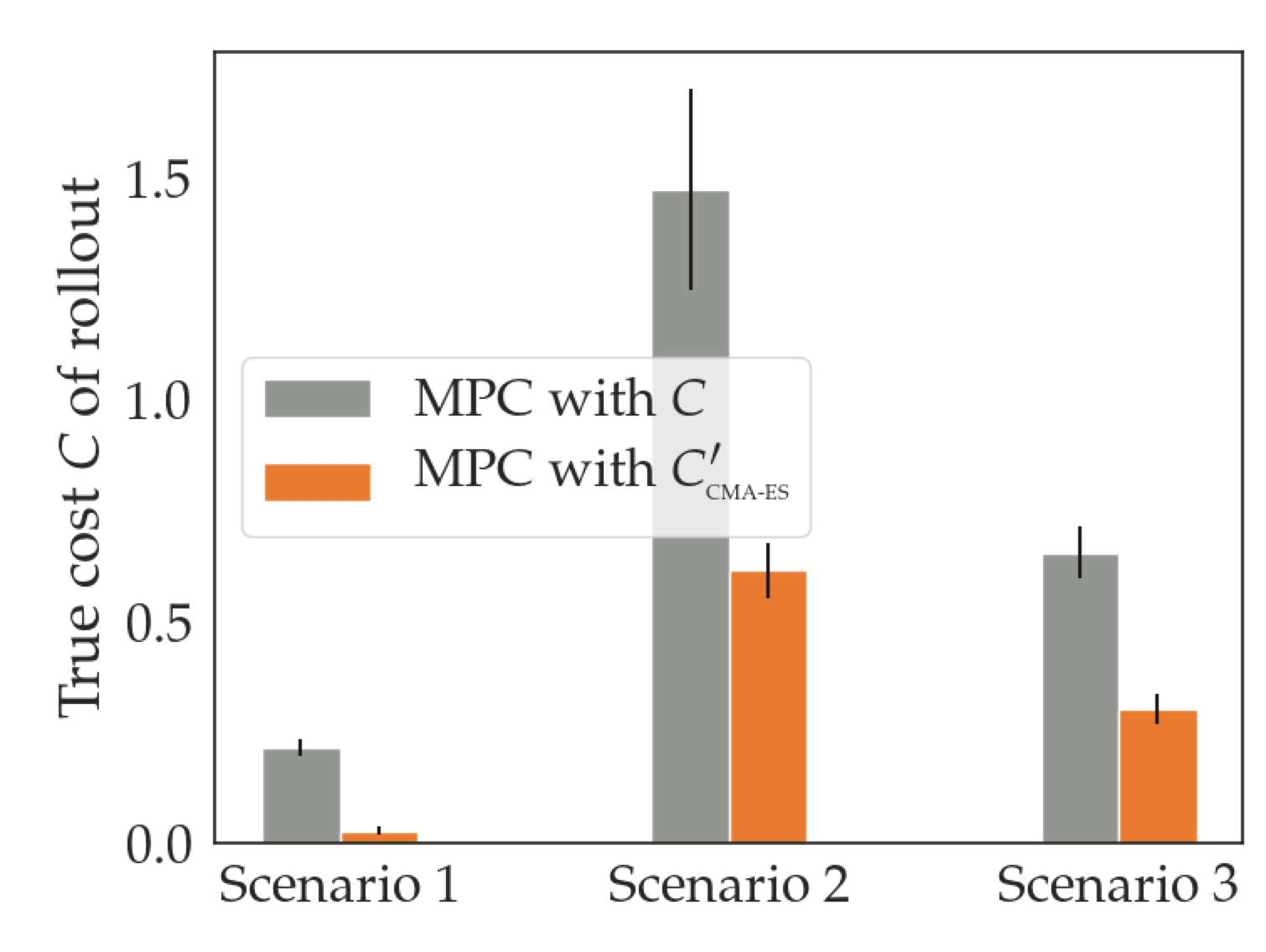}
    }
    \caption{Quantitative Costs under $C$ for the different scenarios in \figureref{fig:case_studies} when there is (a) no dynamics mismatch and (b) a mismatch between the true dynamics and the approximate dynamics model used for MPC planning (see \sectionref{sec:dynamics_mismatch}). Error bars represent standard error over 7 trials.}
    \label{fig:delta_perf}
\end{figure}

\subsection{Scenario 1: Suboptimality Due to Short Horizon}
We first provide an example of how OCD-MPC can alleviate problems resulting from MPC using a shorter planning horizon than the true horizon of the problem. 
% optimizing the MPC objective in \equationref{eq:mpc_objective}, where the optimization is with respect to horizon $K<T$, where $T$ is the true horizon of the problem. %As noted earlier, optimizing over a shorter horizon can lead to myopic behaviors under the true cost function. %Note that while we don't also learn a terminal cost here, the addition of a terminal cost can't always solve horizon problems, since the terminal cost will not be equal to the value function---this leaves headroom for a surrogate cost $C'$ to improve the resulting MPC rollout.
We consider the driving scenario shown in \figureref{subfig:scenario_finite} where the robot car is colored orange and there is a fixed speed human car (black).
%, and a MPC robot car (orange).%\dbnote{If these features are the same across scenarios, then we should talk about them earlier when we discuss the simulator and give math equations for the individual features to try and make the paper as self-contained and reproduceable as possible.} 
This true cost function in this scenario encodes costs for collisions, going off road, driving on lane lines, and traveling slower than a target speed. Because the human car ahead of the robot is traveling slower than the robot's target speed, the robot must either switch lanes to maintain speed or slow down and stay in the center lane---the former behavior is preferred and has lower true cost.

% \begin{wrapfigure}{r}{0.5\textwidth}
% %\begin{figure}[t]
%  % Caption and label go in the first argument and the figure contents
%  % go in the second argument
% \floatconts
%   {fig:delta_perf_scenarios}
%   {\caption{Reduction in true cumulative cost from optimizing a surrogate cost function rather than the true cost function for each of the three three scenarios discussed in \sectionref{case-studies}.% In every case, we find that there exists an alternative reward $C'$ such that MPC planning with respect to $C'$ performs better than MPC using $C$, with respect to the \emph{true cost} $C$.
%   }  }
%   {\includegraphics[width=\linewidth]{figs/performance_cost.png}}
%  %\end{figure}
%  \end{wrapfigure}

Optimizing the true cost with MPC yields suboptimal behavior as shown in \figureref{subfig:scenario_finite} (left). The finite horizon planner used by MPC only plans $K=5$ timesteps into the future, during which the car incurs costs for being between on or close to the lane line, but does not plan far enough into the future to realize the benefit of increased speed afforded by a lane change. Thus, the myopic planner plans to slow down and stay in the middle lane. 
 %Our investigation confirmed that this suboptimal behavior is at least partially a result of short horizon planning, as planning with a longer horizon yields desired lane switching behavior. 
 %This short horizon planning reflects a general problem with using an MPC controller as an optimizer for a long horizon cost function. 
 By contrast, \figureref{subfig:scenario_finite} (right) shows that OCD-MPC finds a cost function $C'$ which causes lane switching even with the default short-horizon MPC planner ($K=5$). This yields lower true cumulative cost under $C$, mitigating the effects of short-horizon planning. In \figureref{fig:delta_perf_scenarios} we show the quantitative difference in performance (measured under the true cost function $C$) between MCP rollouts optimized using $C$ and MPC rollouts optimized with the learned cost function $C'$. \figureref{fig:delta_perf_scenarios} shows that, for Scenario 1, OCD-MPC with CMA-ES achieves a 86\% reduction in true cost compared to to directly optimizing the true cost with MPC. We also found that CMA-ES is more effective at decreasing the true rollout cost than the random search baseline.
  \figureref{subfig:scenario_finite} also shows heatmaps for the true and optimized cost functions. We analyzed the learned surrogate cost function, $C'$, and found that, while the true cost provides no lane preference, the learned surrogate cost penalizes the robot for staying in the center of its lane. This explicitly guides the agent to switch lanes, whereas lane switching is only implicitly implied by $C$ via a penalty for slowing down.

\vspace{.1cm}
\noindent \textbf{Learned Terminal Costs:}
As noted in the introduction, it is common to use a learned terminal cost to alleviate finite horizon issues with MPC.  
% An alternative to OCD-MPC for alleviating finite horizon issues is to learning a terminal cost. 
To compare OCD-MPC to this alternative, we learned a terminal cost using POLO \citep{lowrey2018plan} on the true reward $C$. We found that running MPC with $C$ and the learned terminal cost reduces MPC rollout costs by 60\%, compared to not using a terminal cost. However, running MPC with the learned surrogate cost $C'$ plus the learned terminal cost reduces the cost by a full 80\%. This suggests that learning terminal costs and optimal cost design are complimentary: OCD-MPC can still improve performance, even when MPC uses a learned terminal cost. 
% \lcnote{TODO: maybe run polo 7 times and get error bars?}
 
\subsection{Scenario 2: Suboptimality due to Local Optimization}
Next, we consider how OCD-MPC can alleviate problems associated with only planning locally optimal trajectories via MPC.
Consider the scenario shown in \figureref{subfig:scenario_local},
%. Like the previous scenario, this scenario has three lanes and two cars, where the robot planning via MPC is shown in orange and the other agent is shown in black. The same reward features are used; however, 
where the robot car is initialized to the left of the human car and its task is to maneuver into the right lane. The true cost function $C$ penalizes collisions, going off road, being far from the right lane, and traveling slower than the target speed. 
%, which would be desired behavior if the robot car is trying to exit a highway. 
%Unlike the previous scenario, planning to avoid the other car and steer has lower cost even when only considering the planning horizon. 
\figureref{subfig:scenario_local}(left) shows that optimizing $C$ via MPC with a single control initialization converges to a local optima. 
%In this case, the suboptimality comes from MPC only using a small number of initializations to the local planner, making it difficult to find a collision-free path that into the right lane.
% since seeding the MPC planner with additional control initializations yields lower cost trajectories where the car reaches the right lane. 
By contrast, OCD-MPC is able to learn a surrogate cost function $C'$ which causes lane changing behavior without augmenting the number of  MPC control initializations. As shown on the right in \figureref{subfig:scenario_local}, the robot is able to optimize a trajectory using $C'$ that acheives the desired behavior encoded in the true cost $C$: the surrogate cost trajectory has the robot car (orange) slow down and then change lanes. The quantitative difference in performance under the true cost function $C$ is shown in \figureref{fig:delta_perf_scenarios}. For Scenario 2, we found that OCD-MPC results in a 24\% reduction in true cost compared to optimizing the true cost directly with MPC, and outperforms random search.
We also plot heatmaps in \figureref{subfig:scenario_local} representing the true and learned cost functions. The surrogate cost function reduces the penalty for collision and increases the cost of being far from the right lane.
%while the collision cost weight is five times larger than the right lane preference weight in the true weights, the collision cost weight is slightly smaller than the right lane preference weight in the alternate weights. 
This changes the shape of the reward surface so that even with a single control initialization, gradient descent discovers the globally optimal lane switching behavior via vanilla MPC on $C'$. %Here, the problem of locally optimal planning was directly mitigated by reducing the scale of a feature in the cost which causes the local optima to arise in planning.    
%In the locally optimal planning scenario, the collision weight is reduced in learned weights so trajectory optimization can get over the vertical ridge in the cost landscape where the robot sees itself potentially colliding with the human. 
%Thus, the learned surrogate cost mitigates the local optimality issue directly by scaling down a feature which causes a local minima to exist.

\subsection{Scenario 3: Suboptimality due to Replanning}
In our third scenario, we consider suboptimal behavior that results from the fact that MPC plans trajectories without accounting for future replanning.
%considering the additional information that will be gathered in the future and the opportunities it will have to replan based on this future information.
%We note that MPC performs better than open loop planners because it replans at each step, enabling the robot to react to new information visible in its horizon as well as to changes in the cost landscape. 
%However, when optimizing a linear cost function in an environment with a binary random variable revealed at a fixed timestep, the optimal behavior is to take the action which maximizes expected cost; planning this behavior requires optimizing a trajectory which branches into two possible sequences of controls depending on the outcome, and maximizing the expected total cost over the two paths. This is not a standard feature of MPC, and implementing this would require a more complex trajectory optimizer. To apply standard MPC planning to such a stochastic environment in a risk-averse manner, we might optimize for worst case cost, by simply choosing plans which maximize the minimum of the two possible cost functions.
%dHowever, despite replanning at every time step, MPC does not consider contingencies: it takes a step along a full trajectory that was optimized without considering the replanning that will take place and the additional information that will be obtained at a later time. 
We consider the scenario shown in \figureref{subfig:scenario_replanning} in which the robot is between two lanes, approaching a slow human car which is also between lanes. The true cost incentivizes avoiding collisions, avoiding driving off-road, staying in the center of a lane, and maintaining a high speed. The robot knows the human will merge into one of the two lanes in the future, thinks both cases are equally likely, and plans based on expected cost under its belief distribution. %Note that this scenario has different optimal behaviors depending on the human's outcome. 
Because the robot car incurs cost for traveling slowly behind the human car, the optimal behavior is to merge into whichever lane becomes free, once the human chooses a lane. However, MPC sees a human ahead of it in the worst case regardless of which lane it chooses, causing the robot to slow down to avoid a collision and to arbitrarily pick a lane to merge into to avoid driving on the lane line. MPC does not perform contingency planning and cannot foresee that it can maintain a higher speed and will be in a better position to choose a lane in future timesteps.

The learned surrogate cost function results in emergent contingency planning: the robot maintains its starting velocity and then picks the free lane once the human reveals their lane preference. \figureref{fig:delta_perf_scenarios} shows that, for Scenario 3, OCD-MPC results in a 49\% decrease in true cost compared to optimizing the true cost directly with MPC.
\figureref{subfig:scenario_replanning} shows heatmaps for $C$ and $C'$. The surrogate cost function reduces the lane line penalty and has a larger penalty for not traveling at the target speed, resulting in the robot continuing behind the human and maintaining its speed. 
The MPC planned trajectory gradient is pushed to one lane by asymmetry after the human starts moving into one of the lanes.
% turning direction and resolves the ambiguity over human lane preference. 
Note that in all the above scenarios there is an optimal long term behavior (e.g. switching lanes, staying centered until human steers) which is implicit in the true cost $C$ but can't be produced by optimizing the true cost via MPC. The learned surrogate costs encode this optimal behavior more explicitly, making it easier for MPC to find the optimal solution.

\subsection{Suboptimality due to an approximate dynamics model}\label{sec:dynamics_mismatch}
% We have demonstrated that we we can find a surrogate cost $C'$ which provides better behavior with MPC under $C$ with constant initial conditions. 
The above results assume the robot has access to the true dynamics in its planner, as opposed to an approximate dynamics model (that is, $\hat{f} = f$). We now test what happens when this assumption is relaxed. To create a dynamics mismatch we add anisotropic noise to the true dynamics of the simulator. We simulate wind blowing across the highway by adding a small Gaussian distributed force in each of the three scenarios and then reran OCD-MPC on these three modified scenarios. The robot uses the unmodified $\hat{f}$ (deterministic dynamics model with no wind) when performing MPC, but actually moves according to the transition function $f$ where the car is affected by wind. Note that the fitness function for OCD-MPC does roll out trajectories using the true dynamics $f$. We report our results in \figureref{fig:delta_perf_scenarios_dynamics_mismatch}. Not only can OCD-MPC learn a cost function that improves performance in each scenario, in all three scenarios, OCD-MPC is able to learn a cost function that performs similarly to the case without dynamics mismatch. 
% OCD-MPC is able to learn a cost function that compensates for dynamics mismatch. 

\section{Generalization to unseen initial states}
\label{sec:generalization}
\begin{figure*}
    \centering
    \includegraphics[width=\textwidth]{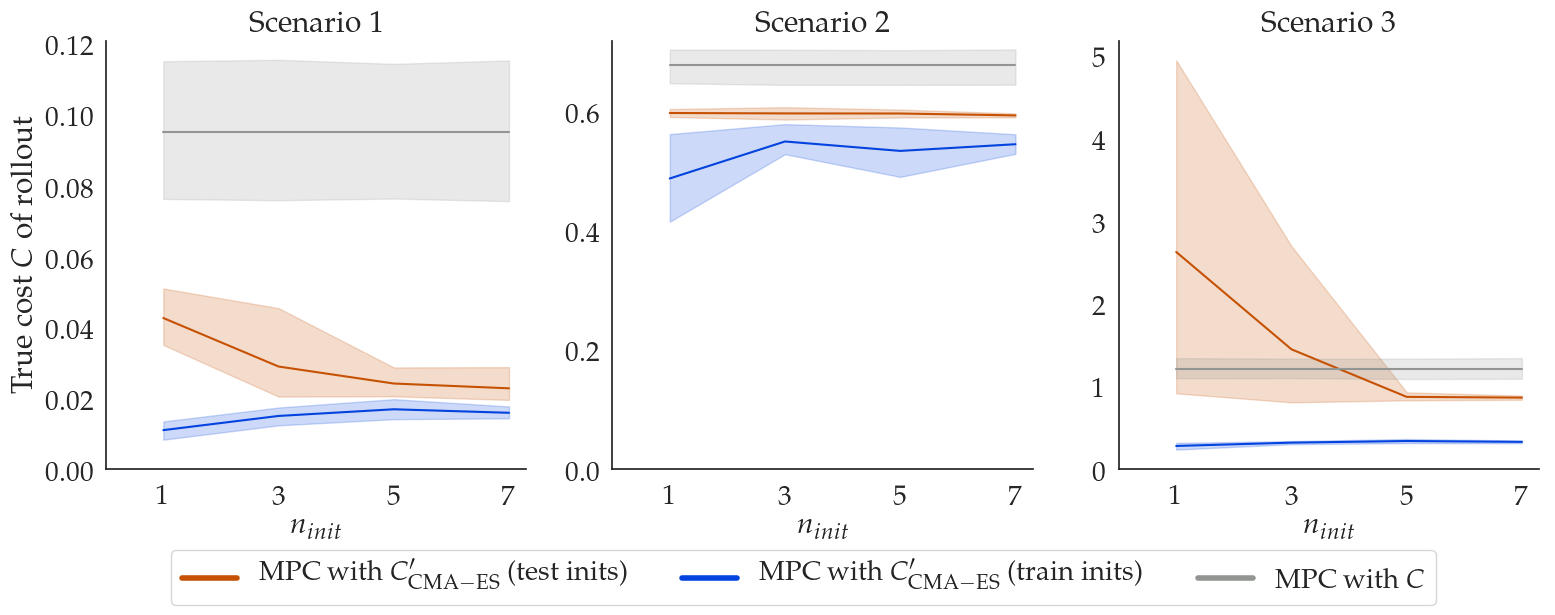}
    \caption{Generalization results from \sectionref{sec:generalization}. In Scenarios 1 and 2, optimizing for a single initialization is sufficient to select a surrogate cost which performs well on unseen state initializations. In Scenario 3 (replanning), we need 5+ initializations to outperform planning with the true cost $C$. Shaded region represents the bootstrapped 95\% confidence interval, calculated over 10 seeds. }
    \label{fig:generalization}
\end{figure*}

The results above demonstrate that OCD-MPC can improve the performance of MPC; 
however, is this because OCD-MPC is overfitting to the single initial start state which we optimize for? 
% Is it possible to find a $C'$ which generally yields lower cost trajectories across similar conditions? 
To test generalization, we varied the initial position and velocity of the car in each of the three scenarios above, and investigate whether we can learn a surrogate cost function that yields lower rollout costs under $C$ across unseen initial conditions. In particular,
%To study the robustness of cost functions learned by OCD-MCP, we evaluate our cost functions under new environment samples for testing. W
we sample new initial states using a Gaussian distribution centered around the initial states shown in \figureref{fig:case_studies}.
%\footnote{Specifically, we varied the initial positions of the cars as well as their speeds.} %DB: I moved this earlier
After sampling $n_{init}$ initial states, we use CMA-ES to solve the optimal reward problem.
%in \equationref{eq:ocd-mpc}. 
We then evaluate the true rollout cost using a test set of 24 new initial states from the same distribution. \figureref{fig:generalization} shows the results over 10 replicates.

% ^Regarding that note, it wouldn't make sense to just provide numbers for stddev right? do I just call them hand-crafted to produce qualitatively similar scenarios ?

% Wouldn't the obvious rejoinder be, the STD is too small? So 
% yes, which maybe is a fair claim on the local opt scenario, but in the others we see real effects on the generalization plots which in itsself should be enough to show the variances are non-negligible.

% Ok save that quote for the rebuttal. 

In Scenario 1 and 2, we find OCD-MPC trained on a single initial condition is sufficient to reduce true rollout cost on unseen testing initial conditions. Furthermore, using more initializations during training yields slightly lower mean cost and lower variance on test initializations, even though this yields a higher cost on the training samples. This suggests that we are able to learn a cost function that generalizes across scenario initializations, while only performing slightly worse in any particular scenario.
%using more samples yields cost functions which may not be able to perform as well on each training sample as a cost functions chosen individually for that sample, using more samples to select a cost function is likely to yield a surrogate cost which is more robust to variations in initial conditions. 
% We observe a benefit to using many training samples, even when the scenario is simple enough that only a single sample needs to be used for consistent improvement over the true weights.
Scenario 3 is more complex, and training on fewer than 5 initializations yields a surrogate cost which occasionally does better than the true cost, but frequently performs worse. However, using more than 5 training initializations yields a surrogate cost which consistently outperforms rollouts generated under the true cost. Scenario 3 reveals the importance of using a training set which provides coverage of the test distribution. %If we don't use enough samples, then the cost function learned by OCD-MCP performs poorly in many unseen conditions. 
% \textbf{TODO} Plot generalization score of best cost weight found by random search vs. number of initial configurations evaluated for each weight in bayesopt/random for locally optimal planning/finite horizon failure cases; we should see generalization score increase.
% \textbf{TODO} Also measure distance between optimized costs and true cost. We expect that if we ask the optimized cost to generalize over more and more tasks (where the objective is in terms of true cost) that the optimized cost will likely converge to the true cost.
%\subsection{Generalization Discussion}
%Our results show that evaluating weights with more initial states for the robot car decreases average generalization cost on unseen initial states. 
To find a $C'$ which is robust to variation in initial state, one should evaluate cost functions on a set of initial states which provides good coverage of the range of possibilities. Furthermore, the existence of scenarios where a surrogate cost function $C'$ can be found which yields lower cost under $C$ and is robust to unseen variations of the scenario suggests that MPC can consistently fails to produce near-optimal behavior even if its cost function accurately encodes true task preferences. %Our true preferences and the cost fed to MPC must be decoupled; we can do better by feeding MPC a cost function which causes it to achieve desired behavior.

\section{Summary and Future Work}\label{sec:conclusion}

% \noindent\textbf{Summary.}
%\noindent\textbf{Summary:} 
In this work, we proposed the Optimal Cost Design problem for MPC (OCD-MPC), as well as a zeroth-order optimization approach to solve this problem. Using this approach, we analyzed three autonomous driving scenarios where optimizing an alternate cost function via MPC yields lower cumulative cost under the true cost function. We also showed a degree of robustness of the learned cost functions to the initial state of our environments. 
%\noindent\textbf{Limitations and Future Work:}
%We used a simple model of uncertainty over human and simple human car preprogrammed behaviors.
% \vspace{.2cm}
% \noindent\textbf{Limitations and future work.}
While we focused on autonomous driving, we believe our approach is general enough to work on a variety of continuous control tasks. 
 Future work includes applying OCD-MPC to more dynamic environments with more nuanced human models and investigating the feasibility of optimizing a single cost function that works well across a wider range of tasks.
% Finally, another direction for future work is 
%better solvers for the OCD-MPC problem. 
%In this work, we used a zeroth-order optimization procedure (CMA-ES). In addition to experimenting with additional zeroth-order optimizers, future work could 
% to approximate the gradient of the cost with respect to the surrogate cost and use standard first-order optimization techniques. 

\section*{Acknowledgements}
This work is partly supported by AFOSR, NSF NRI SCHOOL, and ONR YIP.

\bibliography{references}

% \appendix

% \section{First Appendix}\label{apd:first}

% This is the first appendix.

% \section{Second Appendix}\label{apd:second}

% This is the second appendix.

\end{document}